# Sensorimotor Visual Perception on Embodied System Using Free Energy Principle

Kanako Esaki[1], Tadayuki Matsumura[1], Kiyoto Ito[1] and Hiroyuki Mizuno[1]

[1]Center for Exploratory Research, Research & Development Group, Hitachi, Ltd., Tokyo, Japan
kanako.esaki.oa@hitachi.com

## Abstract

We propose an embodied system based on the free energy principle (FEP) for sensorimotor visual perception. We evaluated it in a character-recognition task using the MNIST dataset. Although the FEP has successfully described a rule that living things obey mathematically and claims that a biological system continues to change its internal models and behaviors to minimize the difference in predicting sensory input, it is not enough to model sensorimotor visual perception. An embodiment of the system is the key to achieving sensorimotor visual perception. The proposed embodied system is configured by a body and memory. The body has an ocular motor system controlling the direction of eye gaze, which means that the eye can only observe a small focused area of the environment. The memory is not photographic, but is a generative model implemented with a variational autoencoder that contains prior knowledge about the environment, and that knowledge is classified. By limiting body and memory abilities and operating according to the FEP, the embodied system repeatedly takes action to obtain the next sensory input based on various potentials of future sensory inputs. In the evaluation, the inference of the environment was represented as an approximate posterior distribution of characters (0–9). As the number of repetitions increased, the attention area moved continuously, gradually reducing the uncertainty of characters. Finally, the probability of the correct character became the highest among the characters. Changing the initial attention position provides a different final distribution, suggesting that the proposed system has a confirmation bias.

## Introduction

The human visual field seems to cover a wide area of the surrounding environment, but the range with high enough resolution to identify the shape and color of the target object is limited to only the central visual field of about 5 degrees (Mandelbaum, J. and Sloan, L. L., 1947). Since the human visual field has this spatial limitation, it is necessary to move the gazing position to obtain environmental information (see) in a wide area. This leads to a phenomenon called sensorimotor contingency (hereafter called SMC; O'Regan, J. K. and Noë, A., 2001; Seth, A.K., 2015), which changes the interpretation of "seeing" (Land, M. F., 2006; Friston, K. and Kiebel, S., 2009; Seth, A.K. et al., 2012; Adams, R. A. et al., 2013; Bogacz, R., 2017; Lotter, W., 2017). Human sensorimotor visual perception makes "seeing" knowing about things to do rather than making an internal representation. In other words, experience is not something we feel but something we do (O'Regan, J. K., 2001). This is because human sensorimotor visual perception infers the environmental state, not the sensory input itself. Therefore, the spatial limitation of the human visual field is not a "limitation" but is actually a "trigger" for an action that moves the gazing position.

In machine learning, many methods have been proposed that incorporate various characteristics and phenomena found in humans. The spatial limitations of the human body, such as that described above, are treated as constraints called partial observation in the context of reinforcement learning (Pineau, J. et al., 2003; Ji, S. et al., 2007; Silver, D. and Veness, J., 2010). These papers evaluated the degree of performance degradation by partial observation in classification and regression problems, assuming that not all necessary information can be obtained in most practical cases. In terms of classification and regression performance, the spatial limitations are certainly constraints and not treated as "triggers" for actions. Various models and algorithms similar to SMC have also been proposed. Auto regressive models (Gregor, K. et al., 2015; Oord, A. et al., 2016; Salimans, T. et al., 2017) predict the entire image by repeating the action of obtaining a partial image. Algorithms that generate exploring actions for reinforcement learning (Oh, J. et al., 2015; Houthooft, R. et al., 2016) contribute to finding optimum solutions by covering a wide range of the action space without bias. These models and algorithms generate the actions not on a basis of inference of environmental states but on that of sensory inputs such as partial images and observations in reinforcement learning. None of the above studies have achieved the inference of environmental states, which is the essence of human sensorimotor visual perception.

The purpose of this study is to achieve human sensorimotor visual perception using the free energy principle (hereafter called FEP; Friston, K. et al., 2006; Friston, K. J., 2010; McGregor, S. et al., 2015; Friston, K. et al., 2017; Buckley, C. L. et al., 2017). The FEP mathematically describes the principle that living things obey. The free energy in the FEP measures the difference between the probability distribution of environmental states that act on a biological system and an approximate posterior distribution of environmental states encoded by the configuration of that system. The biological system minimizes the free energy by changing its configuration to affect the way it samples the environment or by changing the approximate posterior distribution it encodes. These two changes correspond to action and perception, respectively, and lead to "active inference" and "perceptual inference." The biological system thus encodes an implicit and probabilistic model of the environment.

Although the FEP has successfully described a biological system of performing active and perceptual inferences mathematically, it is not enough to model human sensorimotor visual perception. An embodiment of this system is the key to achieving human sensorimotor visual perception. The embodiment provides an interaction between the biological system and environment, resulting in sensory inputs and actions, and the causal relationship between these is stored in the biological system (Fitzpatrick, P. et al., 2003; Cheng, G. et al., 2006; Friston, K., 2011; Gallagher, S. and Allen, M., 2018).

In this paper, we propose an embodied system based on FEP to achieve sensorimotor visual perception. The proposed embodied system is configured by a body, which partially observes the environment, and memory, which has classified prior knowledge about the environment as a generative model. Evaluation using the MNIST dataset (LeCun, Y. and Cortes, C., 1998) shows that the proposed system triggers an action that moves a gazing position and repeatedly performs active and perceptual inferences by following the FEP. Moreover, the intentionality is reproduced on the proposed system, producing an equivalent of human confirmation bias. We discuss how important this bias is for taking the action in an unknown environment.

## Sensorimotor Visual Perception

To list the components necessary for achieving sensorimotor visual perception, the following problem settings are considered. Figure 1 shows an overview of our problem setting. Let us consider a situation in which there is an object to be perceived, e.g., the number 5, in the environment and the vision sensor is used to perceive it. The situation is called an environmental state $x_t$. The vision sensor replaces the role of the human eye and has a spatial limitation of the visual field. The spatial limitation leads to a change in direction of the vision sensor to obtain images in the wide area of the environment. When the direction of the vision sensor is determined in a certain direction, an image of a specific region of the environment is obtained. A representative position of the region (for example, the position of the upper left corner of the region) is defined as an attention position, which equals the gazing position of the human eye. The image of the region is defined as an attention image. The attention image at each time point ($T = t-2, t-1, t$) is obtained by the previous actions $a_{t-3}, a_{t-2}, a_{t-1}$ that move the attention position every time step. A composition image of the attention images obtained up to the current time $T = t$ is used as a sensory input (hereafter called sensory input image $s_t$). As described above, sensorimotor visual perception is to perceive the object (the number 5) in the environment by repeating the actions to obtain $s_t$. Our goal of this study is to achieve sensorimotor visual perception based on FEP.

## Free Energy Principle

The FEP mathematically describes the perceptual and active inferences. In our problem setting, the sensory input image $s_t$ is determined by the previous action $a_{t-1}$ and the current environmental state $x_t$. The $x_t$ does not change ($x_{t-1} = x_t = x_{t+1}$). Under the above assumptions, the variational free energy

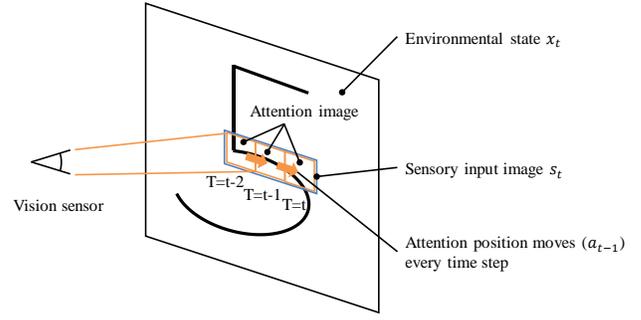

Figure 1 Overview of problem setting to consider sensorimotor visual perception.

and expected free energy, which are necessary components to describe the perceptual and active inferences, are expressed as follows. The variational free energy $F(\phi_{x_t}, a_{t-1})$ described in the FEP is expressed as

$$F(\phi_{x_t}, a_{t-1})$$

$$= E_{q(x_t|\phi_{x_t})}[\ln q(x_t|\phi_{x_t}) - \ln p_{a_{t-1}}(x_t, s_t)]$$

$$= D_{KL}[q(x_t|\phi_{x_t})||p_{a_{t-1}}(x_t|s_t)] - \ln p_{a_{t-1}}(s_t), \quad (1)$$

where $q(x_t|\phi_{x_t})$ is the approximate posterior distribution of $x_t$, and $p_{a_{t-1}}(x_t, s_t)$ is a generative model that stores the causal relationship of $x_t$ and $s_t$ under $a_{t-1}$. The $q(x_t|\phi_{x_t})$ is made to minimize the variational free energy. The expected free energy $G(\phi_{x_{t+1}}, a_t)$ described in the FEP is expressed as

$$G(\phi_{x_{t+1}}, a_t)$$

$$= E_{q(s_{t+1}|x_{t+1}, \phi_{x_{t+1}})} F(\phi_{x_{t+1}}, a_t)$$

$$= E_{q(s_{t+1}|x_{t+1}, \phi_{x_{t+1}})} E_{q(x_{t+1}|\phi_{x_{t+1}})}$$

$$[\ln q(x_{t+1}|\phi_{x_{t+1}}) - \ln p_{a_t}(x_{t+1}, s_{t+1})]$$

$$= E_{q(s_{t+1}|x_{t+1}, \phi_{x_{t+1}})} \Big[ E_{q(x_{t+1}|\phi_{x_{t+1}})}$$

$$[\ln q(x_{t+1}|\phi_{x_{t+1}}) - \ln p(x_{t+1})]$$

$$- E_{q(x_{t+1}|\phi_{x_{t+1}})} \ln p_{a_t}(s_{t+1}|x_{t+1}) \Big]$$

$$= E_{q(s_{t+1}|x_{t+1}, \phi_{x_{t+1}})} D_{KL}[q(x_{t+1}|\phi_{x_{t+1}})||p(x_{t+1})]$$

$$+ E_{q(s_{t+1}|x_{t+1}, \phi_{x_{t+1}})} \Big[ -E_{q(x_{t+1}|\phi_{x_{t+1}})} \ln p_{a_t}(s_{t+1}|x_{t+1}) \Big]. \quad (2)$$

$p_{a_t}(x_{t+1}, s_{t+1})$ is factorized into $p_{a_t}(s_{t+1}|x_{t+1})p(x_{t+1})$ since $x_{t+1}$ and $a_t$ are independent. The next action is selected to minimize the expected free energy.

The purpose of perceptual inference is to find $q(x_t|\phi_{x_t})$ that minimizes $F(\phi_{x_t}, a_{t-1})$. Since the second term of Equation (1) is composed of $a_{t-1}$ and the current $s_t$, which have already been determined, the purpose is achieved by changing $q(x_t|\phi_{x_t})$ with $\phi_{x_t}$ to make the first term close to zero. When the first term is made close to zero, $p_{a_{t-1}}(x_t|s_t) \sim q(x_t|\phi_{x_t})$. Active inference aims to find action

$a_t$ that minimizes $G(\phi_{x_{t+1}}, a_t)$. Since $a_t$ is included only in the second term (hereafter called uncertainty), the purpose is achieved by minimizing the uncertainty.

## Proposed Embodied System for Sensorimotor Visual Perception

An embodiment is the key to achieving sensorimotor visual perception based on the FEP. The proposed embodied system is configured by a body and memory. The body has an ocular motor system for controlling the attention position. In our problem setting shown in Figure 1, the vision sensor is the body. The body has thus limited spatial observation ability, which means that it can only observe a small area of the environment that is focused upon. The memory is not photographic but is a generative model that contains prior knowledge about the environment, and that knowledge is classified. By limiting body and memory abilities and operating according to the FEP, the proposed embodied system repeatedly performs perceptual and active inference. Figure 2 shows the processing flow of sensorimotor visual perception with the proposed system. In perceptual inference, a current sensory input image $s_t$ is generated using an attention image obtained from the environment and the past attention images obtained up to the current time point. Then, $s_t$ is input to the generative model $p_{a_{t-1}}(x_t, s_t)$ to calculate an approximate posterior distribution $q(x_t|\phi_{x_t})$ and an approximate posterior image $q_{img}(x_t|\phi_{x_t})$ which is a conversion of $q(x_t|\phi_{x_t})$ into an image format. In active inference, the uncertainty of the expected sensory input images $s_{t+1}$ is calculated using $q_{img}(x_t|\phi_{x_t})$. After that, an attention position having the minimum value is selected from an uncertainty map in which uncertainty is mapped for each $s_{t+1}$. In generating $s_t$ (the first process of perceptual inference), the attention image obtained from the environment and the past attention images are composed while maintaining their relative attention positions to generate $s_t$. In calculating $q(x_t|\phi_{x_t})$ and $q_{img}(x_t|\phi_{x_t})$ (the second process of perceptual inference), $p_{a_{t-1}}(x_t, s_t)$ is implemented with a combination of a variational autoencoder (VAE) and a fully connected neural network (FNN), which is inspired by the idea of auxiliary loss in GoogleNet (Szegedy, C. et al., 2015). The VAE contains prior knowledge about the environment, and the FNN classifies the prior knowledge in the latent space generated in the VAE. The combination of the VAE and FNN is pre-trained, which corresponds to human "experience." The pre-training is equivalent to changing $\phi_{x_t}$ in Equation (1) to make the first term of that equation closer to zero. Although $\phi_{x_t}$ is updated in the long term, it is fixed in the short term from $t-1$ to $t$. The $q(x_t|\phi_{x_t})$ and $q_{img}(x_t|\phi_{x_t})$ are calculated by inputting the $s_t$ to the combination of the VAE and FNN. Calculation of uncertainty (the first process of active inference) uses $q_{img}(x_t|\phi_{x_t})$. Uncertainty is the expected information amount of $p_{a_t}(s_{t+1}|x_{t+1})$, which is the probability distribution of $s_{t+1}$ with action $a_t$ as a parameter, which is conditioned by $x_{t+1}$. Conditioning by $x_{t+1}(= x_t)$ is interpreted as extracting $s_{t+1}$ from $q_{img}(x_t|\phi_{x_t})$. The $s_{t+1}$ is composed of the current $s_t$ and the candidate attention images surrounding it. Parameterizing $a_t$ is interpreted as assuming the candidate attention images. Under these interpretations, the information amount of $p_{a_t}(s_{t+1}|x_{t+1})$ is calculated as that of $s_{t+1}$. The information amount of $s_{t+1}$ is the entropy of the approximate posterior distribution calculated by inputting $s_{t+1}$ to $p_{a_{t-1}}(x_t, s_t)$. Since $s_{t+1}$ is deterministically calculated from one $q_{img}(x_t|\phi_{x_t})$, the expected-value calculation is not required. Uncertainty is thus the set of the entropies of each $s_{t+1}$. In selecting the attention position (the second process of active inference), an uncertainty map is generated where the entropy of each $s_{t+1}$ corresponds to each attention position. An attention position that minimizes the entropy is selected according to the uncertainty map.

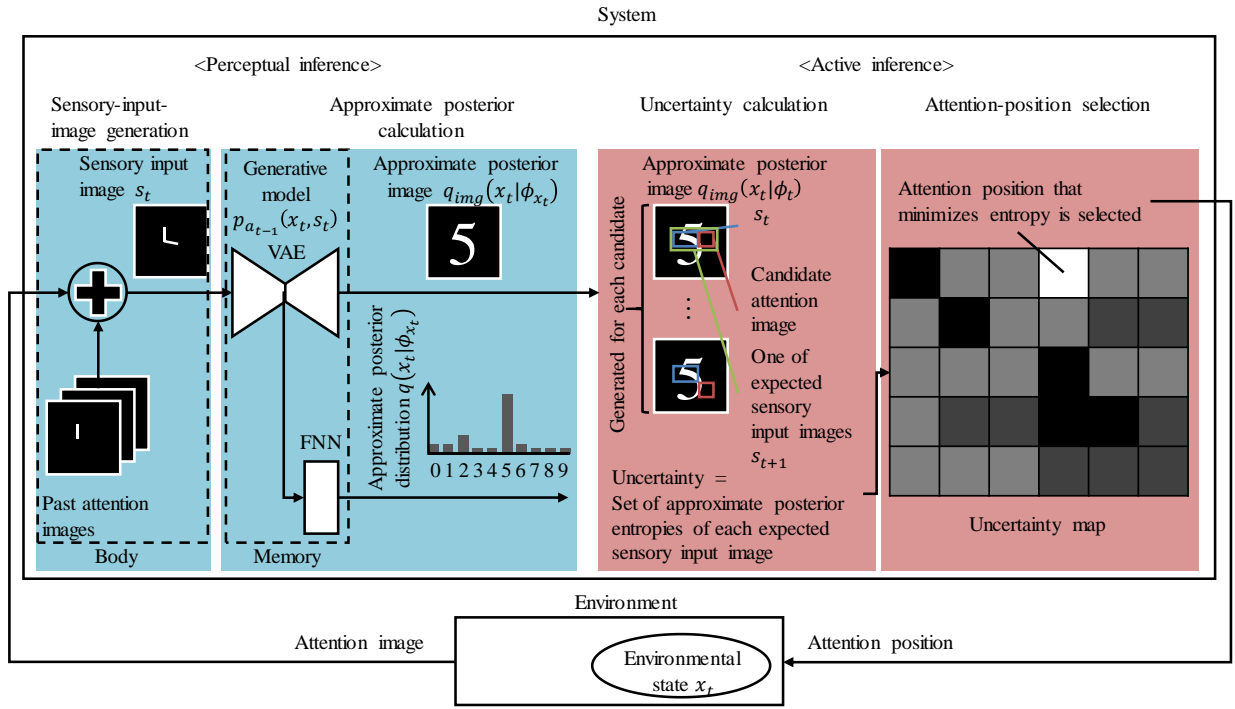

Figure 2 Processing flow of sensorimotor visual perception with proposed embodied system.

Algorithm 1 shows the pseudo code of the processing flow. The $p_{a_t}(s_t, x_t)$ is pre-trained using training data of $(s_t, x_t)$. All the training data of $s_t$ are pre-processed so that the center of gravity of an image is shifted to the center position. During the operation of the proposed embodied system, the process from the 2nd line to the 12th line is repeated. First, an attention image $s'_t$ is obtained from the vision sensor. The past sensory input images are composed with the obtained $s'_t$ while maintaining each relative attention position. The center of gravity of the composed image is calculated, and the composed image is shifted so that the center of gravity is located at the center position of the image. The shifted composed image is an $s_t$. Then, $q(x_t|\phi_{x_t})$ is calculated by inputting $s_t$ to $p_{a_t}(s_t, x_t)$. After that, the sub function starting from the 14th line is called to generate expected sensory input images $s_{t+1}$. In the sub function, an $q_{img}(x_t|\phi_{x_t})$ is calculated by inputting $s_t$ to $p_{a_t}(s_t, x_t)$. A template image is generated by detecting a bounding rectangle area of non-zero pixels in $s_t$ and extracting the area from $s_t$. Template matching is carried out in the $q_{img}(x_t|\phi_{x_t})$, and the representative position of the current $s_t$, $u_{cur}$, is obtained. To calculate the next candidate attention positions $u_{next}$, a candidate region of $u_{next}$ is set. The candidate region is a region obtained by adding a fixed margin pixel to a region of $s_t$ in $q_{img}(x_t|\phi_{x_t})$. The region of $s_t$ is defined by $u_{cur}$ and the size of the template image. The $u_{next}$ are calculated by sliding the window with the fixed stride pixel in the candidate region. The window is the size of $s'_t$. The representative positions of all the window positions during sliding are $u_{next}$. The $s_{t+1}$ are generated by extracting the region of the $s_t$ and the region of the next candidate attention images $s'_{t+1}$ from $q_{img}(x_t|\phi_{x_t})$. The region of the $s_t$ is defined by $u_{cur}$ and the size of the template image, as mentioned above. The region of the $s'_{t+1}$ are defined by $u_{next}$ and the size of $s'_{t+1}$. The extracted images are clipped or applied with zero-padding to have the same size as $s_t$. Each approximate posterior distribution $q(x_{t+1}|\phi_{x_{t+1}})$ is calculated by inputting each image included in $s_{t+1}$ to $p_{a_t}(s_t, x_t)$. Note that $q(x_t|\phi_{x_t})$ is calculated using the current $s_t$, while $q(x_{t+1}|\phi_{x_{t+1}})$ is calculated using $s_{t+1}$. The entropy of each $q(x_{t+1}|\phi_{x_{t+1}})$ is calculated and added to the uncertainty map $M$. Finally, the attention position having the minimum value in $M$ is defined as the next attention position.

| Algorithm 1 Pseudocode of processing flow |
|---|
| 1:    while system is operating do: |
| 2:       obtain attention image $s'_t$ |
| 3:       compose past sensory input images with $s'_t$ |
| 4:       generate sensory input image $s_t$ by shifting gravity point to center of composed image |
| 5:       calculate approximate posterior distribution $q(x_t|\phi_{x_t})$ using $s_t$ and generative model $p_{a_t}(s_t, x_t)$ |
| 6:       call generate_expected_sensory_input_image |
| 7:       for expected sensory input images $s_{t+1}$: |
| 8:           calculate approximate posterior distribution $q(x_{t+1}|\phi_{x_{t+1}})$ using $s_{t+1}[index]$ and $p_{a_t}(s_t, x_t)$ |
| 9:           calculate entropy of $q(x_{t+1}|\phi_{x_{t+1}})$ |
| 10:          add entropy to uncertainty map $M$ |
| 11:       end for |
| 12:       set attention position using $M$ |
| 13:    end while |
| 14:    generate_expected_sensory_input_image |
| 15:       generate approximate posterior image $q_{ima}(x_t|\phi_{x_t})$ using $s_t$ and $p_{a_t}(s_t, x_t)$ |
| 16:       generate template from $s_t$ |
| 17:       obtain current sensory input position $u_{cur}$ by template matching in $q_{ima}(x_t|\phi_{x_t})$ |
| 18:       calculate next candidate attention positions $u_{next}$ using $u_{cur}$ |
| 19:       generate $s_{t+1}$ using $q_{img}(x_t|\phi_{x_t})$, $u_{cur}$, and $u_{next}$ |
| 20       return with $s_{t+1}$ |

## Evaluation

We evaluated the proposed embodied system for achieving sensorimotor visual perception in a character-recognition task with partial observation constraint. We used the MNIST dataset as the character dataset. The generative model was implemented with a convolutional VAE (Kingma, D.P. and Welling, M., 2014; Rezende, D. J. et al., 2014) that was robust against displacement and a three-layer FNN (30 units in the hidden layer, 10 units in the output layer, and the activation function of the output layer was the Softmax function). Training of the generative model involved 60,000 MNIST images (28 × 28 pixels). The stochastic variables $x_t$ of the approximate posterior distribution $q(x_t|\phi_{x_t})$ were labels 0–9 of the MNIST dataset. The size of both the sensory input image and approximate posterior image was 28 × 28 pixels, while the size of the attention image was 6 × 6 pixels. The margin pixel and stride pixel for generating the next candidate attention images were set to 3 and 1, respectively. The number of attention repetitions was set to 20.

Perceptual inference was performed during the attention repetitions. Figure 3 shows the transition of $q(x_t|\phi_{x_t})$ when making an attention at characters "0", "2", "9", and "3". Symbols A, B, C, and D in Figure 3 (a) are described later. Each graph plots the probability distribution for the stochastic variable (0–9). In each subfigure ((a)–(d)), the graph on the left shows the distribution for the 1st to 10th attentions in order from the top, and the graph on the right shows the distribution for the 11th to 20th attentions in order from the top. The image on the right side of each graph is the corresponding approximate posterior image $q_{img}(x_t|\phi_{x_t})$. Regardless of the characters, the probability of "1" was maximum from the 1st to the 2nd attention, and the $q_{img}(x_t|\phi_{x_t})$ at that time contained a shape that was difficult to recognize as a character. After the 3rd attention, the probability of the characters different from the correct characters "0", "2", "9", and "3" changed to the maximum, and the $q_{img}(x_t|\phi_{x_t})$ at that time contained shapes similar to numbers "5" and "7". When making an attention at characters "0", "2", and "9", the probability of correct

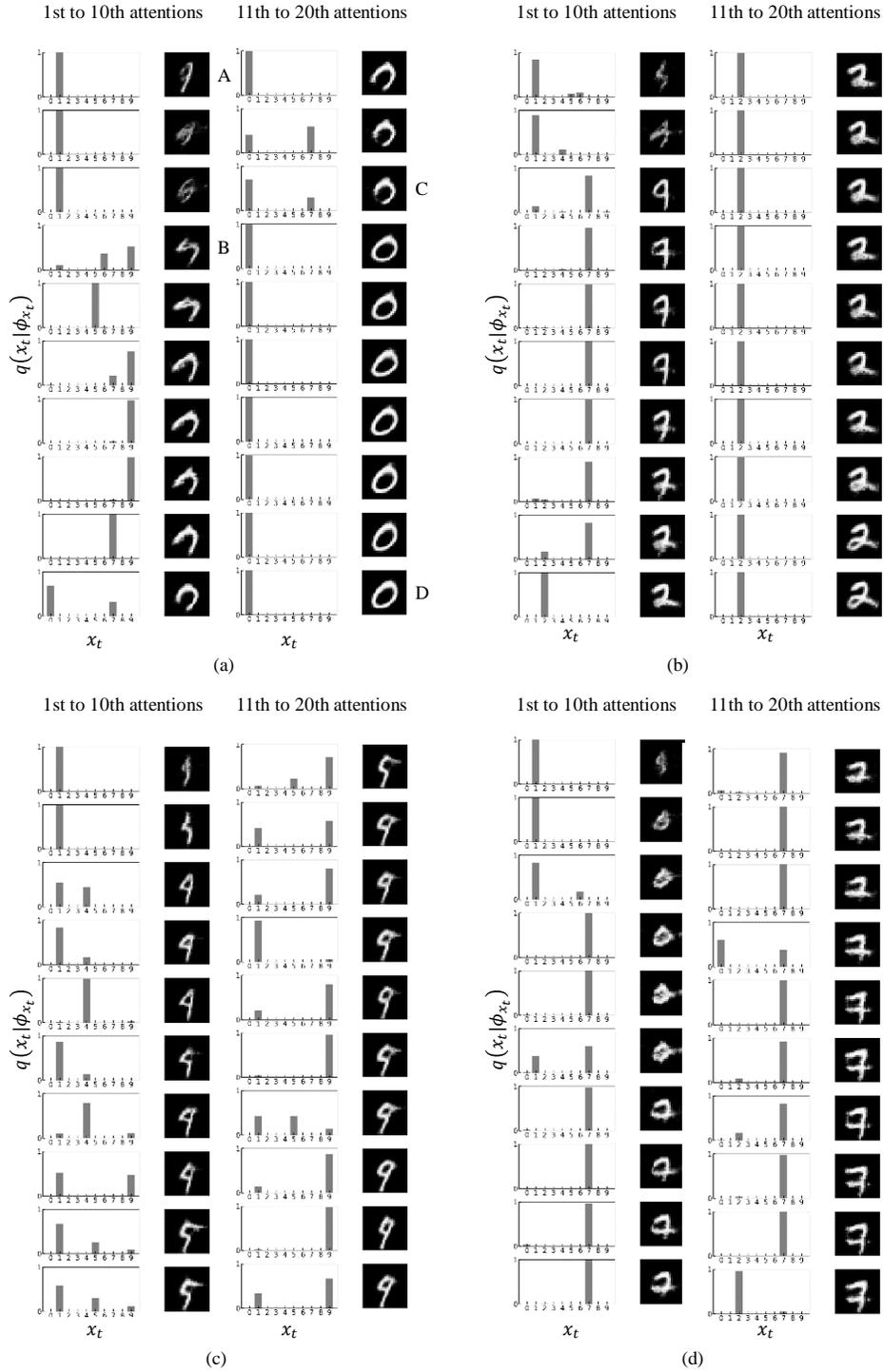

Figure 3 Transition of approximate posterior distribution along with attention repetition. Each subfigure (a), (b), (c), and (d) corresponds to making attention at the characters "0", "2", "9", and "3", respectively.

characters was maximum at the 20th attention, and "0", "2", and "9" appeared in the $q_{img}(x_t|\phi_{x_t})$. On the other hand, when making an attention at character "3", the probability of "7" was maximum from the 4th to 13th and from the 15th to 19th attentions, and the probability of "2" was maximum even at the 20th attention. The $q_{img}(x_t|\phi_{x_t})$ contained a part of "3", not the entire "3".

Active inference was performed during attention repetitions. Figure 4 shows the uncertainty map at the 1st, 4th, 13th, and 20th attentions (corresponding to A, B, C, and D in Figure 3 (a), respectively) when making an attention at "0". The vertical (x) and horizontal (y) axes of the maps indicate the position coordinates of the upper left corner of the candidate attention image. The values of the uncertainty map of the 1st and 20th attentions were biased toward the minimum value, whereas the values of the uncertainty map of the 4th and 13th attentions were dispersed.

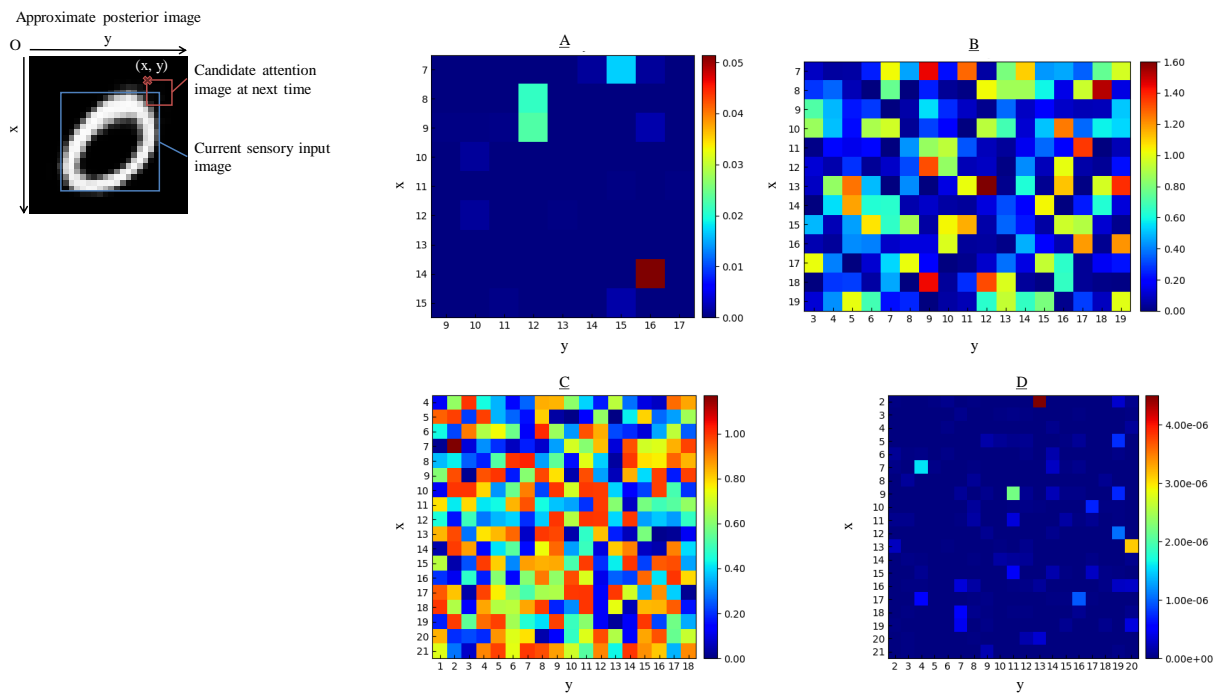

Figure 4 Uncertainty map when making attention at "0". Each subfigure A, B, C, and D corresponds to 1st, 4th, 13th, and 20th attentions, respectively.

Sensorimotor visual perception was performed by repeating perceptual and active inferences. Figure 5 shows attention images and sensory input images for each number of attentions when making attentions from "0" to "9". Each subset of two rows corresponds to making attentions from "0" to "9". The first row shows the attention images (The rectangle area of each image is the attention image). The second row shows the sensory input images. As the number of attention repetition increased, the visible area of a character became larger, and at the 20th attention, except "3", most of the characters were visible enough to be recognized by humans as numbers. At the 20th attention of "3", however, a part of "3" was missing.

To analyze the case of making an attention at "3", the initial attention position was changed. Figure 6 shows transition of approximate posterior distribution, attention image, and sensory input image when the initial attention position was changed to a position different from those in Figures 3 and 5 in the case of making an attention at "3". Similar to Figure 3 (d), the probability of "1" was maximum from the 1st to 4th attentions, and the approximate posterior image at that time contained shapes that were difficult to recognize as characters. Different from Figure 3 (d), the probabilities of "1", "3", and "5" changed to maximum after the 5th attention, and the probability of correct character "3" reached maximum at the 20th attention. The probability image contained "3".

## Discussion

In the transition of the approximate posterior distribution, the probability of "1" was maximun at the 1st attention for all characters from "0" to "9" that were verified. This is because, at the 1st attention, the sensory input image only includes the 1st attention image capturing only points or short lines, and this is perceived as a part of "1". In fact, the approximate posterior images also looked like "1." This is similar to the situation in which humans temporarily labeled information (in this case, "1") so that they could identify the environment based on the information they had collected and their experience thus far and moved on to the next action.

In the middle stage of attention repetition for most of the characters, a form that could not be identified as any shape was shown in the approximate posterior image; accordingly, the approximate posterior distribution fluctuated among various shapes during attention repetition. This situation is analogous to a human making a decision without a specific classification label, considering the various possibilities when information is uncertain.

While most characters fluctuated among various shapes in the middle stage of the attention, the period in which "7" was higher was maintained in the middle stage of the attention repetition when making the attention of "2". This is because the shape of "2" includes the shape of "7", and "7" appears first in this attention-position order. In fact, the approximate posterior image in the middle stage of the attention repetition was "7".

In the early stages of attention repetition, the probability of a particular number in the approximate posterior distribution will not increase wherever the next attention is made. The distribution of the uncertainty map is thus biased toward the minimum value of the uncertainty. In this case, the difference in the value included in the uncertainty map is mainly due to the random number included in the VAE calculation and considered to be in the situation equivalent to random selection of the next attention position. This is similar to a situation in which a human takes a random attention action to obtain clues in an unknown environment.

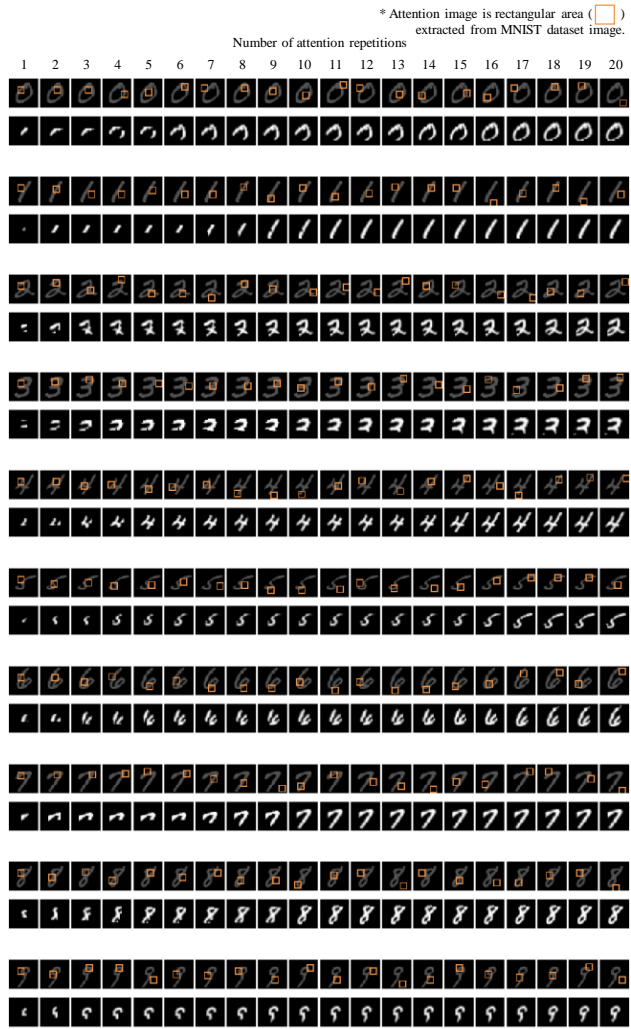

Figure 5 Transition of attention images and sensory input images. Each subset of two rows corresponds to making attentions from "0" to "9". First row shows attention images* and second row shows sensory input images along with number of attention repetitions.

In the middle stage of the attention repetition, the number can be specified depending on the next sensory input image determined with attention position, so the distribution of the uncertainty map is more dispersed than that in the early stage of attention repetition. This is equivalent to a human taking actions to increase the certainty of a certain number when the likelihood of that number increases.

In the final stage of attention repetition, most regions of numbers have already become visible in the approximate posterior image; thus, the distribution of the uncertainty map is again biased toward the minimum-value side. This is a situation in which the probability of a particular number in the approximate posterior distribution has already been high, and the entropy of this distribution is low no matter where the next attention position is. For this reason, the range was smaller than that of the uncertainty map in the early stage of attention repetition in which the distribution was similarly biased.

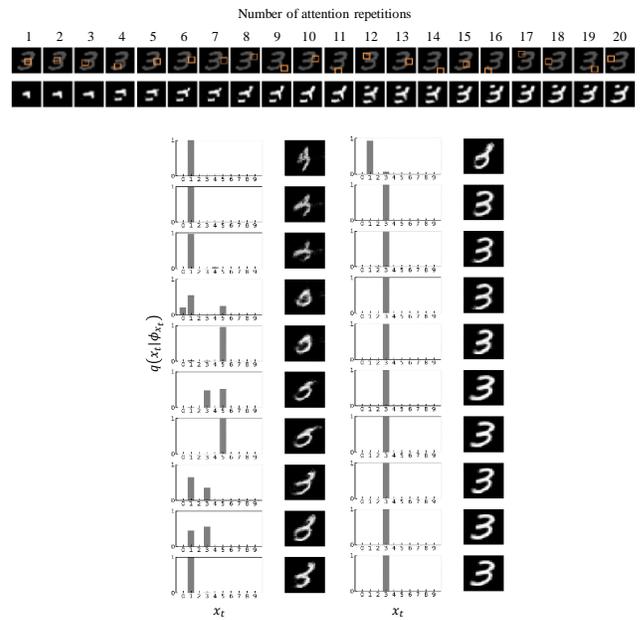

Figure 6 Attention images, sensory input images, and transition of approximate posterior distribution. Changes in initial attention position when making attention at "3" provides different transition patterns.

In the transition of attention images, the attention-position movement was along the line of the character, but not a complete "one-stroke". This is because the approximate posterior image is not necessarily the image with the correct number. A number different from the correct number is shown in the approximate posterior image from the early to middle stages of attention repetition, and the uncertainty is calculated using the approximate posterior image. Even at the final stage of the attention repetition, due to the randomness of the VAE, a random movement can also be seen along the character line.

When making an attention at "3", different initial attention positions confirmed both cases in which the probability of correct characters became the highest and that of different characters became the highest after a fixed number of attention repetitions. This result suggests that the proposed system has a confirmation bias similar to humans that depends on what is obtained from the environment and prior knowledge about the environment. The human confirmation bias has a negative impact on decision-making in various fields, from politics to science and education (Kappes, A., 2020). Whereas, the confirmation bias has the advantage of adaptability to unknown environments. General optimization problems require complete models of the environment, but lack of environmental information and time constraints cannot provide complete models in most practical cases. In these cases, humans take the next action on the basis of the confirmation bias made of previous experiences. Taking the next action (exploring action) enables making decisions in practical time and obtaining the environmental information. We believe that our results will help solve the previously difficult problem of triggering action in an unknown environment.

## Conclusion

We proposed an embodied system based on the free energy principle (FEP) for sensorimotor visual perception. We evaluated the proposed system in a character-recognition task using the MNIST dataset. The proposed embodied system is configured by a body and memory. By limiting body and memory abilities and operating according to the FEP, the proposed system triggers an action that moves an attention position and repeatedly performs perceptual and active inferences. In the evaluation, as the number of repetitions increases, the attention area moves continuously, gradually reducing the uncertainty of the characters. Finally, the probability of the correct character becomes the highest among the characters. It was thus confirmed that the embodied system greatly contributes to achieving sensorimotor visual perception. Moreover, changing the initial attention position provides a different final distribution, suggesting that the proposed system has a confirmation bias similar to humans. We believe that these results will help solve the difficult problem of triggering action in an unknown environment.